\documentclass[10pt,twocolumn,letterpaper]{article}

\usepackage[pagenumbers]{wacv}
\usepackage[accsupp]{axessibility}

\definecolor{wacvblue}{rgb}{0.21,0.49,0.74}
\usepackage[pagebackref,breaklinks,colorlinks,allcolors=wacvblue]{hyperref}

\def\wacvPaperID{99} %
\def\confName{WACV}
\def\confYear{2027}

\title{Where Did the Light Come From? Diagnosing Cast-Shadow Reasoning in VLMs and Borrowing the Missing Shadow Prior from Image Generation Models}

\author{
Ryugo Morita$^{1,2}$ \quad Shunsuke Kitada$^{3}$ \quad Stanislav Frolov$^{1}$ \quad Brian Bernhard Moser$^{1}$\\
Ko Watanabe$^{1}$ \quad Riku Takahashi$^{1,3}$ \quad Jinjia Zhou$^{3}$ \quad Andreas Dengel$^{1,2}$\\
{\small $^{1}$DFKI, Germany \quad $^{2}$RPTU Kaiserslautern-Landau, Germany \quad $^{3}$Hosei University, Japan}
}

\begin{document}

\twocolumn[{
\maketitle
\begin{center}
\includegraphics[width=0.93\textwidth]{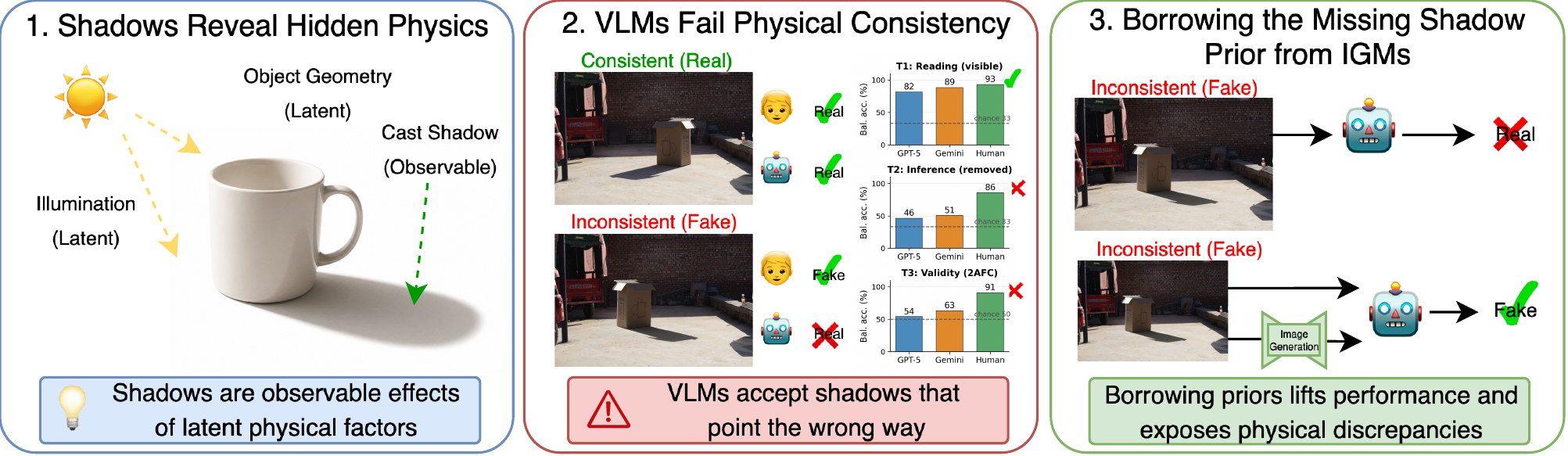}
\par\vspace{8pt}
\refstepcounter{figure}
\label{fig:teaser}
\parbox{\textwidth}{
\small
{Figure~\thefigure.}
We introduce \textbf{ShadowBench}, which probes one facet of illumination understanding, whether VLMs know which way a cast shadow should fall. (i)~\textbf{Reading}: VLMs read a \emph{visible} shadow's direction well above chance, scaling with size. (ii)~\textbf{Inferring}: with the shadow removed, they recover it weakly. (iii)~\textbf{Validating}: they barely exceed chance on a bias-free two-alternative test, while humans reach $85$--$98\%$. Letting the VLM borrow an Image Generation Model's cast shadow prior lifts its weak inference and exposes wrong-direction shadows it otherwise accepts. The prior VLMs lack already resides in IGMs, pointing toward transferring it into VLM training.}
\end{center}
}]

\begin{abstract}
Shadows are observable consequences of latent physical causes, arising from the interaction between illumination, object geometry, and scene structure. A model that understands shadows should be able not only to \emph{read} a shadow's direction, but also to \emph{infer} where a shadow should fall when it is absent, and to \emph{judge} whether a shadow is physically consistent with the scene. We introduce \textbf{ShadowBench}, built from real photographs, which poses these three questions to vision-language models (VLMs). From each scene, we build a correct-shadow, a shadow-removed, and a human-annotated \emph{wrong-direction} state. Evaluating $19$ open VLMs ($2$B to $78$B) and frontier proprietary models, we find that VLMs read a visible shadow's direction well above chance and improve with scale, yet infer the absent direction only weakly and fall to chance at judging physical consistency even under a bias-free two-alternative test that removes the yes-bias of single-image naturalness questions. Only Gemini-3.1-pro exceeds chance on validity, and even then only modestly and far below humans. Chain-of-Thought (CoT) does not help. The missing ingredient is not reasoning but knowledge that image generation models (IGMs) already hold. Routing a VLM through an IGM lifts its inference well above its direct attempt and helps it flag the wrong-direction shadows it otherwise accepts. Cast-shadow direction is a single facet of illumination understanding, but the sharp gap it reveals between VLMs and IGMs suggests image-generation or reconstruction objectives as a route toward illumination-aware VLMs.
\end{abstract}

\section{Introduction}
\label{sec:intro}
A cast shadow is a rare gift for studying physical understanding, a single visible mark that encodes one hidden variable, the direction of the light. A shadow that points the wrong way makes an ordinary photograph look unmistakably broken~\cite{debevec2008rendering}, making shadows an unusually clean probe of whether a model reasons about scene physics or merely about pixels. Shadows have long been exploited through detection~\cite{masquil2025s}, removal~\cite{xu2025detail}, and generation~\cite{wang2025metashadow}, and modern diffusion models~\cite{rombach2022high,esser2024scaling} control illumination explicitly~\cite{kocsis2024lightit,zhang2025scaling}.
Yet as vision-language models (VLMs) are promoted to general visual reasoners, a basic question remains unanswered. \textit{Do they understand the physics a shadow encodes, or only register that a shadow is there?}

Producing or accepting a plausible shadow does not require understanding it. A cast shadow encodes essentially one physical quantity, the direction of the light relative to the object, and a model that has internalized this relation should be able to use it in three ways, \emph{(i)~reading} the direction of a visible shadow, \emph{(ii)~inferring} where a shadow should fall when it is absent, and \emph{(iii)~validating} whether a shown shadow is physically consistent with the scene. Reading only locates a visible cue, whereas inferring and validating require relating the shadow to the scene's global information. Existing physical-reasoning benchmarks probe spatial relations~\cite{kamath2023s,chen2024spatialvlm}, 3D structure~\cite{wang2024picture,rong2026unicbench}, occlusion~\cite{liu2025beyond}, and broad plausibility~\cite{chow2025physbench, wang2025forensics}, but none isolate cast-shadow direction, and none separate \emph{reading} a shadow from \emph{understanding} it, which is precisely the distinction that decides whether a model has internalized the shadow--illumination relation.

To this end, we introduce \textbf{ShadowBench}, built from real photographs. From each scene, we recover the object and its shadow as an alpha-matted layer, and recompose three matched states on a shared background, a shadow-correct, a shadow-removed, and a human-curated wrong-direction state. Evaluating 19 open VLMs ($2$B--$78$B) and frontier proprietary models, we find a sharp split. Reading works; VLMs read a visible shadow's direction well above chance, improving with scale. Understanding does not; the same models infer the absent direction only weakly and cannot reliably tell a wrong-direction shadow from a correct one, accepting it as natural even under a bias-free two-alternative test, while humans perform every task at $85$--$98\%$. The sole exception, Gemini-3.1-pro, exceeds chance only modestly. Chain-of-Thought (CoT)~\cite{kojima2022large} repairs neither failure, so the bottleneck is not reasoning. 

VLMs are trained to produce text conditioned on images, with no objective requiring them to reproduce the physics of a scene, whereas Image Generation Models (IGMs) are trained to reconstruct physically plausible images and must respect illumination, geometry, and the other constraints the visible world obeys. We therefore let the VLM borrow this knowledge through a loop it drives end-to-end, prompting an IGM to draw the shadow the scene implies and then reading it back, rather than judging from the scene alone. This recovers both inference and validity well above the VLM's direct attempt, supplying the missing prior that CoT did not. It points to folding generation or reconstruction objectives into VLM training as a route to internalizing the priors IGMs hold. Our contributions are summarized as follows:

\begin{itemize}
\item We introduce \textbf{ShadowBench}, which uses cast shadows to probe whether VLMs hold the illumination-direction knowledge a cast shadow encodes. Matched alpha-composites (shadow-correct, shadow-removed, and a wrong-direction state) make direction the only changed variable, and three balanced tasks separate \emph{reading} a visible shadow from \emph{inferring} an absent one and \emph{validating} a shown one, each with a human baseline.

\item Across 19 open VLMs ($2$B--$78$B) and frontier proprietary models, VLMs read a visible shadow's direction well above chance but infer the absent direction weakly and, with the sole exception of Gemini-3.1-pro, stay at chance when telling a wrong-direction shadow from a correct one, even under CoT, while humans reach $85$--$98\%$.

\item We show that the cast-shadow knowledge VLMs lack already resides in IGMs, and that bringing these illumination priors into the VLM's decision lifts its inference and flags many of the wrong-direction shadows it otherwise accepts, motivating generation-based training and knowledge transfer toward illumination-aware VLMs.
\end{itemize}

\section{Related Works}
\subsection{Illumination Control and Estimation}
Illumination-aware image modeling has evolved from harmonization, which aligns a composite's foreground with its background~\cite{li2024freepih, ren2024relightful, yu2024cfdiffusion}, through increasingly fine-grained relighting~\cite{ponglertnapakorn2023difareli, kim2024switchlight, choi2025scribblelight}, to diffusion models that condition explicitly on shading, radiance, or shadow cues~\cite{zhang2025scaling, kocsis2024lightit, zeng2024dilightnet, fortier2024spotlight, zhou2024foreground}. Generative models thus encode strong, controllable priors over light and shadow, but these priors are used to \emph{render} illumination into an image, not to read it back out. Conversely, inverse rendering and lighting estimation recover scene illumination explicitly~\cite{zhu2022irisformer, lin2025iris, dai2025inverse}; these methods are \emph{trained} to invert image formation, so given suitable inputs they solve our inference task by construction. ShadowBench asks a different question, whether \emph{general-purpose} VLMs hold this knowledge at all. The informative comparison is VLMs vs.\ IGMs, as neither is trained for illumination estimation, yet only the latter holds a usable directional prior (Sec.~\ref{subsec:image}). This locates the gap in the training objective rather than in the task's solvability.

\begin{figure*}[t]
\centering
\includegraphics[width=0.97\textwidth]{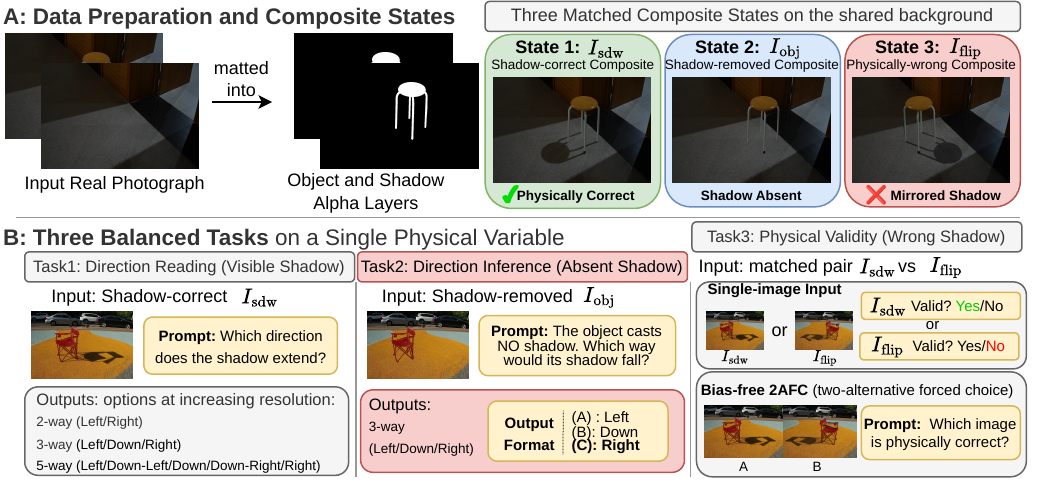}
\caption{Overview of \textbf{ShadowBench}. From each real photograph, we recover the object and its cast shadow as an alpha-matted layer and recompose matched states on a shared background: the shadow-correct $I_{\mathrm{sdw}}$, the shadow-removed $I_{\mathrm{obj}}$, and the physically-wrong composite $I_{\mathrm{flip}}$, produced by a horizontal flip to change the shadow's direction. The target object is highlighted in red. These states drive three tasks of increasing demand on the shadow--illumination relation: \textbf{Direction Reading} from a visible shadow at increasing angular resolution (Task~1); \textbf{Direction Inference}, recovering the direction from $I_{\mathrm{obj}}$ once the shadow is removed (Task~2); and \textbf{Physical Validity}, judging whether a shown shadow is consistent with the scene, in single-image and a bias-free two-alternative form on $I_{\mathrm{sdw}}$ vs.\ $I_{\mathrm{flip}}$ (Task~3).}
\label{fig:bench}
\end{figure*}

\subsection{Shadow Modeling and Analysis}
Shadows have long served computer vision as cues for scene understanding and realism~\cite{debevec2008rendering}. Work on detection~\cite{masquil2025s} and removal~\cite{xu2025detail} showed that robust shadow handling needs global context and object--shadow associations rather than local appearance alone~\cite{guo2023shadowformer}, and recent work extends to generation and composition~\cite{liu2024shadow, zhou2024foreground, wang2025metashadow, hou2024compose, cai2025parametric}. These works operate at the pixel level, treating the shadow itself as the target rather than as a cue to the single latent variable that produced it, the direction of the light.

\subsection{Benchmarks for Vision-Language Models}
VLM evaluation~\cite{bai2025qwen3,zhu2025internvl3,hong2025glm} has expanded from recognition to structured reasoning: spatial relations~\cite{kamath2023s,chen2024spatialvlm}, 3D and metric understanding~\cite{wang2024picture,rong2026unicbench}, occlusion~\cite{liu2025beyond}, physical plausibility~\cite{chow2025physbench, wang2025forensics}, and causal or counterfactual probes under controlled interventions~\cite{komanduri2025causalvlbench, zhang2025cf}. No benchmark isolates shadow as a physical variable or distinguishes reading a visible shadow from inferring or validating an absent one; ShadowBench fills this gap with matched interventions that change only the shadow's direction relative to the scene.

\section{ShadowBench}
\label{sec:shadowbench}
As shown in Fig.~\ref{fig:bench}, ShadowBench is built from the real-image dataset collected with OBER~\cite{zhao2025objectclear}, which provides, for each scene, a clean background image $I_{\mathrm{back}}$, an image $I_{\mathrm{all}}$ captured after placing a foreground object, and annotated masks for the object ($M_{\mathrm{obj}}$) and for the region affected by its cast shadow ($M_{\mathrm{eff}}$).

We isolate the cast shadow as a soft alpha layer. Within the shadow region $R = M_{\mathrm{eff}} \setminus M_{\mathrm{obj}}$, a shadowed pixel is darker than the clean background, so its opacity follows from how far the per-pixel ratio between the two captures drops below one,
\begin{equation}
\alpha = \mathrm{clip}\!\left(1 - \mathrm{lum}\!\left(I_{\mathrm{all}} \oslash I_{\mathrm{back}}\right),\, 0,\, 1\right) \odot \mathbb{1}[R],
\end{equation}
where $\oslash$ is per-pixel division and $\mathrm{lum}(\cdot)$ denotes luminance. The clip maps non-shadowed pixels, whose ratio meets or exceeds one, to zero opacity. Writing the object as an alpha-matted layer $L_{\mathrm{obj}} = (C_{\mathrm{obj}}, M_{\mathrm{obj}})$ and the compositing operator $\mathrm{Comp}(B, (C,A)) = A \odot C + (1-A) \odot B$, we recompose three matched states on the shared background $I_{\mathrm{back}}$:
\begin{align}
I_{\mathrm{sdw}}  &= \mathrm{Comp}\!\left(I_{\mathrm{back}} \odot (1-\alpha),\; L_{\mathrm{obj}}\right), \\
I_{\mathrm{obj}}  &= \mathrm{Comp}\!\left(I_{\mathrm{back}},\; L_{\mathrm{obj}}\right), \\
I_{\mathrm{flip}} &= \mathrm{Comp}\!\left(I_{\mathrm{back}} \odot (1-\alpha^{\leftrightarrow}),\; L_{\mathrm{obj}}^{\leftrightarrow}\right),
\end{align}
where $(\cdot)^{\leftrightarrow}$ denotes a horizontal mirror of the foreground (object and shadow), applied while the background is held fixed. $I_{\mathrm{sdw}}$ is the physically-correct composite with the object and its cast shadow; $I_{\mathrm{obj}}$ removes the shadow; and $I_{\mathrm{flip}}$ keeps the identical object and shadow but mirrors them as a unit, so the shadow points inconsistently with the unchanged scene lighting. The mirror is applied only to the foreground, so the background's own illumination cues (neighboring shadows, sky, lighting) are shared by $I_{\mathrm{sdw}}$ and $I_{\mathrm{flip}}$, and the two composites differ only in the direction of the shadow relative to this fixed lighting. Table~\ref{tab:canonical_states} summarizes the three states and the tasks each one drives.

\begin{table}[t]
\centering
\small
\begin{tabular}{llccl}
\toprule
State & Source & Shadow & Phys. & Used in \\
\midrule
$I_{\mathrm{sdw}}$  & composite  & correct   & correct & Task~1,3 \\
$I_{\mathrm{obj}}$  & composite  & removed   & wrong     & Task~2,3 \\
$I_{\mathrm{flip}}$ & composite  & mirrored  & wrong   & Task~3 \\
\bottomrule
\end{tabular}
\caption{Canonical scene states in ShadowBench. $I_{\mathrm{sdw}}$ serves as the direction-\emph{reading} stimulus and the validity positive, $I_{\mathrm{flip}}$ as the wrong-direction negative with the foreground mirrored, and $I_{\mathrm{obj}}$ as the \emph{inference} input and the negative of the presence 2AFC.}
\label{tab:canonical_states}

\end{table}

\subsection{Curation and ground truth}
\label{sec:curation}
The raw pool of composites contains failure cases that would confound the tasks, and the direction tasks require a per-scene ground-truth direction. We curate in two stages.

\noindent\textbf{Stage 1: Background consistency.} We automatically reject scenes whose residual $|I_{\mathrm{all}} - I_{\mathrm{back}}|$ is large outside the object and shadow regions, since a large off-object residual signals a confound between the two captures (camera motion, background motion, or a lighting change) that would corrupt the extracted layers.

\noindent\textbf{Stage 2: Human filtering and direction ground truth.} Three annotators independently inspect each surviving scene and reject it if the object is incompletely separated, the object mask is defective (clipped, fragmented, or spilling onto the background), or the underlying photograph is unusable (heavy motion blur or severe over/under-exposure). A scene is excluded if \emph{any} annotator marks \emph{any} failure. For each accepted scene, the same annotators place a single point at the center of the object's cast shadow at its base, and we average these points across the three annotators. The shadow's own center is taken as the centroid of the shadow mask, and the per-scene ground-truth direction is the angle of the line from the averaged origin point to that centroid. From the $2{,}715$ scenes, curation yields $458$ scenes with reliable per-scene direction ground truth for the reading and inference tasks (Tasks~1 and 2).

For the validity task (Task~3), the mirrored foreground is kept at its flipped position by default. Only when this position is physically implausible (e.g.\ the object floats or intersects scene geometry) does an annotator shift the foreground to the nearest plausible location, applying the smallest correction needed. Curation counts for every stage are listed in Appendix Sec.~\ref{sec:appendix_curation}.

\subsection{Three tasks on a single physical variable}
\label{sec:three_tasks}
As shown in Fig.~\ref{fig:bench}, ShadowBench probes the shadow direction implied by the scene through three tasks. \textbf{Task~1} (Direction Reading, Sec.~\ref{sec:task1}) asks the model to read the direction of a \emph{visible} cast shadow; \textbf{Task~2} (Direction Inference, Sec.~\ref{sec:task2}) removes the shadow and asks the model to infer that direction from the remaining scene; and \textbf{Task~3} (Physical Validity, Sec.~\ref{sec:task3}) asks the model to judge whether a shown shadow is physically consistent with the scene, using the matched composites $I_{\mathrm{sdw}}$, $I_{\mathrm{flip}}$, and $I_{\mathrm{obj}}$. Reading needs only to locate a visible cue, whereas inference and validity both require relating the shadow to the rest of the scene, including other cast shadows, directional shading, and visible light sources. All three tasks are posed as multiple-choice questions using a fixed set of directional words, scored against the per-scene ground-truth directions.

\subsubsection{Task 1: Direction Reading}
\label{sec:task1}
The first task asks whether the model can read the direction of a \emph{visible} cast shadow. Given the shadow-correct image $I_{\mathrm{sdw}}$ with the target object highlighted, the model chooses the direction in which the object's shadow extends. To probe the granularity of this ability, we pose the question at three angular resolutions: a two-way left/right choice (chance $50\%$), a three-way left/down/right choice (chance $33\%$), and a five-way choice left/down-left/down/down-right/right (chance $20\%$). We omit the upward directions, which never occur in the captured scenes for ground-cast shadows. Reading sets the ceiling against which the two comprehension tasks below are compared. We read from $I_{\mathrm{sdw}}$ rather than the original capture $I_{\mathrm{all}}$ to match the compositing pipeline of the other tasks, and verify the two give the same reading accuracy (Appendix Sec.~\ref{sec:appendix_readsource}).

\subsubsection{Task 2: Direction Inference}
\label{sec:task2}
The second task removes the shadow entirely and asks the model to infer where it would fall. Given the shadow-removed composite $I_{\mathrm{obj}}$ with the target object highlighted, the model chooses the direction in which the object's shadow would extend if it were present, as a three-way left/down/right choice (chance $33\%$). With the shadow absent, the answer can only be recovered from the scene (neighboring shadows, sky, lighting), so this task assesses whether the model can infer the latent direction rather than read it directly. It uses the same option set and ground-truth direction as the three-way form of Task~1, making the two directly comparable, so the gap between them isolates the cost of removing the visible shadow.

\subsubsection{Task 3: Physical Validity}
\label{sec:task3}
The third task asks whether the model can judge if a shadow is physically consistent with the scene, which again requires relating the shadow to the rest of the scene rather than merely locating it. We use the matched composites $I_{\mathrm{sdw}}$ (correct) and $I_{\mathrm{flip}}$ (wrong-direction) in both single-image and bias-free two-alternative forms.

\noindent\textbf{Single-image (Yes/No).} The model is shown one image, $I_{\mathrm{sdw}}$ or $I_{\mathrm{flip}}$, and asked whether the highlighted object's shadow is physically consistent. The correct answer is Yes for $I_{\mathrm{sdw}}$ and No for $I_{\mathrm{flip}}$. This form mirrors the naturalness questions common in prior work.

\noindent\textbf{Bias-free two-alternative (2AFC).} To remove the response bias such a single-image question can invite, we place $I_{\mathrm{sdw}}$ and $I_{\mathrm{flip}}$ side by side and ask which is physically correct. Every pair is presented twice, with the correct image on each side, so any fixed-option strategy scores 
$50\%$ and preference separates from discrimination. Both images contain a shadow that differs only in direction, so the flip isolates direction as the only cue separating the correct composite from the wrong one. We call this the \emph{direction} 2AFC. As a control, we also include a \emph{presence} 2AFC that pits $I_{\mathrm{sdw}}$ against the shadow-removed $I_{\mathrm{obj}}$, testing only whether the model expects a grounded object to cast \emph{some} shadow. Contrasting these two 2AFCs separates a model's sensitivity to shadow direction from its sensitivity to shadow presence.

\section{Experiment}
\subsection{Experiment Setup}
\noindent
\textbf{Dataset and metrics.}
We evaluate ShadowBench on $458$ curated scenes, each with a reliable per-scene ground-truth direction. Direction Reading poses each scene at three angular resolutions (two-, three-, and five-way), and Direction Inference at three-way resolution. Physical Validity runs in two forms, a single-image Yes/No over both states ($916$ balanced records) and a within-scene counterbalanced 2AFC presenting each pair in both orders ($458$ pairs, $916$ records). Every task is scored by \emph{balanced accuracy}, the mean of per-class recall, and compared against a human baseline.

\noindent
\textbf{Evaluated models.}
We evaluate $19$ open VLMs from $2$B to $78$B across five families: the Qwen-VL series (2/2.5/3-VL~\cite{bai2025qwen3}), InternVL3~\cite{zhu2025internvl3}, LLaVA-OneVision~\cite{li2024llava}, Gemma-4~\cite{team2024gemma}, and Pixtral~\cite{agrawal2024pixtral}, together with two frontier proprietary models, GPT-5 and Gemini-3.1-pro. For the human baseline, three master's students distinct from the curation annotators each answer the same multiple-choice questions on a random subset of $80$ scenes per task, and we report their mean accuracy. The transfer analysis routes a single editor, FLUX.1-Kontext-dev~\cite{labs2025flux}, and we compare it against three further image generation models, the open OmniGen2~\cite{wu2025omnigen2} and the proprietary gpt-image-2 and Gemini-3-pro-image (Appendix Sec.~\ref{sec:appendix_editor_choice}). All proprietary models are accessed via their official APIs in June 2026.

\begin{table}[t]
\centering
\footnotesize
\setlength{\tabcolsep}{4pt}
\newcommand{\grouprow}[1]{%
  \midrule
  \addlinespace[0pt]
  \multicolumn{7}{c}{\itshape #1}\\
  \addlinespace[-3pt]
  \midrule
}
\begin{tabular}{l@{\hspace{5pt}}ccccccc}
\toprule
& \multicolumn{3}{c}{Reading (T1)} & Infer (T2) & \multicolumn{2}{c}{Validity (T3)} \\
\cmidrule(lr){2-4}\cmidrule(lr){5-5}\cmidrule(lr){6-7}
Model & 2w & 3w & 5w & 3w & Y/N & 2AFC \\
\grouprow{$\leq 4$B (2--4B)}
Qwen2-VL-2B    & 50.0 & 33.3 & 20.9 & 28.5 & 50.0 & 50.0 \\
Qwen3-VL-2B    & 68.4 & 35.7 & 18.1 & 33.0 & 50.0 & 50.0 \\
Qwen2.5-VL-3B  & 50.0 & 47.2 & 27.3 & 33.8 & 50.3 & 50.0 \\
Gemma-4-E4B    & 69.7 & 61.8 & 33.7 & 34.4 & 49.7 & 50.0 \\
\grouprow{7--9B}
Qwen2-VL-7B    & 72.0 & 71.6 & 42.2 & 41.7 & 50.6 & 51.3 \\
Qwen2.5-VL-7B  & 78.8 & 48.1 & 42.3 & 33.3 & 51.1 & 50.2 \\
InternVL3-8B   & 71.6 & 55.9 & 33.9 & 33.3 & 50.0 & $11.9^\dagger$ \\
Qwen3-VL-8B    & 80.0 & 66.7 & 44.1 & 33.5 & 50.2 & 50.2 \\
\grouprow{12--14B}
Gemma-4-12B    & 86.9 & 85.8 & 52.6 & 42.0 & 50.0 & 56.0 \\
Pixtral-12B    & 77.0 & 45.6 & 22.0 & 34.2 & 50.6 & 53.4 \\
InternVL3-14B  & 72.2 & 60.4 & 23.7 & 34.5 & 50.2 & 49.8 \\
\grouprow{26--38B}
Gemma-4-26B    & 88.2 & 86.6 & 57.3 & 43.8 & 52.5 & 51.9 \\
Qwen3-VL-30B   & 76.9 & 63.1 & 32.2 & 37.0 & 51.4 & 52.5 \\
Gemma-4-31B    & 90.0 & 86.8 & 67.4 & 42.0 & 51.3 & 50.3 \\
Qwen2.5-VL-32B & 78.2 & 57.9 & 38.1 & 34.2 & 50.3 & 50.0 \\
InternVL3-38B  & 81.6 & 58.4 & 24.5 & 33.6 & 51.6 & 51.4 \\
\grouprow{72--78B}
Qwen2.5-VL-72B & 79.1 & 75.8 & 38.1 & 37.5 & 50.8 & 55.2 \\
LLaVA-OV-72B   & 80.5 & 62.3 & 31.0 & 33.9 & 51.3 & 54.7 \\
InternVL3-78B  & 68.1 & 59.2 & 31.0 & 32.7 & 50.5 & 48.6 \\
\grouprow{Proprietary}
GPT-5          & 88.1 & 81.7 & 52.2 & 46.4 & \textbf{52.7} & 54.1 \\
Gemini-3.1-pro & \textbf{92.4} & \textbf{88.6} & \textbf{74.8} & \textbf{50.8} & 52.4 & \textbf{63.0} \\
\midrule
\textit{Human}  & \textit{97.6} & \textit{93.2} & \textit{84.7} & \textit{85.7} & \textit{92.6} & \textit{90.9} \\
\textit{Chance} & 50.0 & 33.3 & 20.0 & 33.3 & 50.0 & 50.0 \\
\bottomrule
\end{tabular}
\caption{Main results on \textbf{ShadowBench} (balanced accuracy, \%), all $19$ open and $2$ proprietary models. \emph{Reading} (T1) is at two-/three-/five-way resolution; \emph{Infer} (T2) is three-way with the shadow removed; \emph{Validity} (T3) is single-image Yes/No and the bias-free 2AFC. $\dagger$: verbose non-parseable answers, scored as incorrect. Reading is well above chance and scales with size, while inference stays near chance and validity is at chance for every model except Gemini-3.1-pro ($63.0\%$), far below humans.}
\label{tab:main}

\end{table}

\subsection{Results \& Analysis}
\label{sec:results}

\subsubsection{T1: Reading a visible shadow}
Table~\ref{tab:main} (Reading columns) shows that VLMs read the direction of a \emph{visible} cast shadow well above chance, and that the ability improves with model size (Spearman $\rho=0.53$ over the open models). The trend is cleanest within a single family, where Gemma-4 rises $70\!\rightarrow\!87\!\rightarrow\!90\%$ from $4$B to $31$B. In two-way left/right reading, the accuracy climbs from $50$--$70\%$ for sub-$4$B models to $88$--$92\%$ for the strongest models. Gemini-3.1-pro reaches $92.4\%$ and Gemma-4-31B $90.0\%$, approaching the human $97.6\%$. Size does not fully determine the score (InternVL3-78B reaches only $68\%$), but every model is well above chance. Finer angular resolution lowers every score yet preserves the ordering, with three-way reading peaking at $88.6\%$ and five-way at $74.8\%$, both from Gemini-3.1-pro, against human $93.2\%$ and $84.7\%$.
Two Google-lineage models stand out here, Gemini-3.1-pro leading every resolution, and the larger open Gemma-4 models surpassing the frontier GPT-5 at every resolution.
When the shadow is present, it is a cue that most models can locate and verbalize, the smallest (sub-$4$B Qwen) staying at chance. The gap to humans widens with angular resolution, but reading is not where the capable models fail.

\begin{table}[t]
\centering
\footnotesize
\setlength{\tabcolsep}{3pt}
\resizebox{\columnwidth}{!}{%
\begin{tabular}{l ccc cc ccc}
\toprule
 & \multicolumn{3}{c}{Direction 3-way} & \multicolumn{2}{c}{Validity Y/N} & \multicolumn{3}{c}{Validity 2AFC} \\
\cmidrule(lr){2-4}\cmidrule(lr){5-6}\cmidrule(lr){7-9}
Model & Read & Infer & top & bal & top & bal & top & both$\checkmark$ \\
\midrule
Gemma-4-12B    & 85.8 & 42.0 & R\,75 & 50.0 & N\,100 & 56.0 & B\,52 & 29.2 \\
Gemma-4-26B    & 86.6 & 43.8 & R\,84 & 52.5 & N\,90 & 51.9 & B\,68 & 18.8 \\
Qwen3-VL-30B   & 63.1 & 37.0 & L\,81 & 51.4 & Y\,94 & 52.5 & B\,93 & 9.4 \\
Gemma-4-31B    & 86.8 & 42.0 & R\,87 & 51.3 & N\,50 & 50.3 & B\,98 & 2.2 \\
InternVL3-38B  & 58.4 & 33.6 & R\,98 & 51.6 & Y\,76 & 51.4 & B\,93 & 7.5 \\
Qwen2.5-VL-72B & 75.8 & 37.5 & R\,66 & 50.8 & Y\,97 & 55.2 & B\,61 & 23.8 \\
\midrule
GPT-5          & 81.7 & 46.4 & R\,57 & 52.7 & Y\,92 & 54.1 & B\,85 & 17.2 \\
Gemini-3.1-pro & 88.6 & 50.8 & R\,48 & 52.4 & N\,77 & 63.0 & A\,61 & 45.5 \\
\bottomrule
\end{tabular}
}
\caption{The read-to-infer dissociation, per class. \emph{Read}/\emph{Infer} are three-way balanced accuracy with the shadow present/removed; \emph{top} is the most-predicted class and its share (for validity: Yes/No, or side A/B); \emph{both$\checkmark$} is the share of pairs answered correctly under both 2AFC orders, where a fixed-side strategy scores $0\%$ and random guessing $25\%$. Once the shadow is removed, inference collapses onto a single direction (\emph{top} $57$--$98\%$) and balanced accuracy falls toward chance; single-image validity collapses to a fixed Yes/No and the 2AFC to a fixed side. The full panel is in Table~\ref{tab:perclass_full}.}
\label{tab:perclass}

\end{table}

\subsubsection{T2: Remove the shadow and inference}
The Infer column of Table~\ref{tab:main} tells the opposite story. With the cast shadow removed, every model falls to within a few points of the $33.3\%$ three-way chance. The best model is Gemini-3.1-pro at $50.8\%$, the only entry clearly above chance and still $35$ points below the human $85.7\%$. GPT-5 reaches $46.4\%$ and the strongest open model (Gemma-4-26B) $43.8\%$, while the bulk of the output sits at $33$--$35\%$. As shown in Table~\ref{tab:perclass}, this chance score is not random guessing but a collapse onto a constant answer, with $14$ of the $19$ open models predicting a single direction on at least $80\%$ of scenes, so balanced accuracy becomes one near-perfect class recall averaged with two near-zeros. Gemma-4-26B is the clean case. It reads the three-way direction at $86.6\%$ with the shadow drawn, yet once the shadow is removed its inference falls to $43.8\%$, a $43$-point drop, as it defaults to ``right'' on $84\%$ of scenes (per-class recalls $10/28/93$). The strongest readers fall the furthest, with Gemma-4-31B dropping $86.8\!\rightarrow\!42.0\%$ and Gemma-4-12B $85.8\!\rightarrow\!42.0\%$, so the directions a model reads almost perfectly become near-unrecoverable the moment the shadow is gone. The collapse does not stem from the answer format. Inference shares its prompt and option order with three-way reading, which the same models solve, and shuffling the options changes nothing (Appendix Sec.~\ref{sec:appendix_shuffle}). Reading and inferring a shadow are different abilities, and current VLMs have only the former.

\subsubsection{T3: Judging physical validity}
The Validity columns ask whether a shadow is physically consistent with the scene. In the single-image Yes/No form (Table~\ref{tab:main}), every model is pinned at chance ($49.7$--$52.7\%$), and the per-class breakdown reveals the cause in Table~\ref{tab:perclass}. Most models answer ``yes, consistent'' regardless of the image, a real-photograph default that holds even when the shadow is mirrored to the wrong direction. The default direction is itself model-specific. Most models pin to ``yes'', but the Gemma family and Gemini to ``no'', so the fixed answer reflects a response bias rather than the image, leaving even Gemini-3.1-pro at $52.4\%$. The bias-free 2AFC, which removes this default by forcing a choice between a correct- and a wrong-direction shadow, leaves most of the field unchanged. Every open model stays within $3$ points of $50\%$, and GPT-5 reaches only $54.1\%$. The per-class recall exposes the same image-independent strategy in a new guise. $11$ of the $19$ open models answer a single side (A or B) on $90\%$ or more of trials regardless of which shadow is correct, and the counterbalanced design cancels that constant to exactly $50\%$. Because every pair is shown in both presentation orders, option preference and discrimination separate: a preference-only strategy is never correct on both orders of a pair (random guessing: $25\%$), yet the panel's median both-orders-correct rate is just $4.85\%$ (Appendix Table~\ref{tab:perclass_full}). The constant answers are image-independent, not a preference masking latent sensitivity. Gemini-3.1-pro is the one exception. The same model sits at chance on Yes/No, where its ``no'' bias masks any real signal, yet reaches $63.0\%$ on the 2AFC once that bias is removed, and is correct on both orders of $45.5\%$ of pairs, confirming image-driven discrimination rather than option preference in Table~\ref{tab:perclass}. It is the only model to clear chance on validity, though still far below the human $90.9\%$. Validity is the hardest of the three tasks and, for all but this model, entirely out of reach.

\noindent
\textbf{The validity failure is specific to direction.}
The validity 2AFC pits a correct shadow against a wrong-\emph{direction} one. A failure there has two possible causes. The model might recognize the shadow but misjudge its \emph{direction}, or it might be unable to use the shadow at all. The 2AFC alone cannot tell these apart. We run a second 2AFC, counterbalanced identically, that instead pits the correct-shadow image against the shadow-\emph{removed} image, the same object casting no shadow. The output is far above chance. The open models average $80\%$ (Gemma-4-12B and Qwen2.5-VL-72B reach $93\%$), against $51\%$ on the direction 2AFC in Appendix Table~\ref{tab:presence}. Models thus hold a coarse prior that a grounded object should cast \emph{some} shadow and can flag its absence, but cannot tell whether a shadow that \emph{is} present points the right way. The presence comparison can be partly solved by a low-level ``the more complete image is the real one'' heuristic, whereas the direction 2AFC is immune to it, since both images contain a shadow.

\noindent
\textbf{The chance-level validity is not an artifact of the mirror.}
The mirror changes the shadow's direction by $\geq 90^\circ$ in $85.4\%$ of the
validity scenes (median $153^\circ$), and dropping the $14$ near-vertical scenes
whose direction barely moves leaves every score essentially unchanged
(Appendix Sec.~\ref{sec:appendix_curation}). The mirror also moves the object's shading together with its shadow, and the chance-level result might therefore reflect the foreground rather than the shadow. Rotating the shadow matte alone, with the object pixels untouched, does not settle this. Models read such negatives at $64.0\%$, well above chance, as a rotated shadow is no longer a shape the pictured object could cast, and the contradiction is visible without reading the scene's lighting. We therefore rebuild the negative so that the shadow's shape stays coherent (Appendix Fig.~\ref{fig:editorchain}). The shadow-rotated matte is placed on a white canvas with the scene removed, so the editor has no lighting context to re-infer the true direction from, and the redrawn shadow is clamped to the rotated matte's support so that the direction cannot drift. FLUX.1-Kontext-dev then redraws the shadow, which we composite back into the original photograph before pasting the original foreground on top, leaving the object bit-identical to the positive. Three annotators judge
$98$ of the $1{,}374$ edits free of visible artifacts. On these, models reach $52.3\%$, close to the $51.4\%$ they reach on the mirror. Any artifact the editor leaves behind would make the negative easier to spot, which makes this an upper bound on what the manipulation leaks. We keep the mirror as the benchmark's negative, as a shadow-only rotation carries its own confounds and the editing step succeeds on only some scenes (Appendix Sec.~\ref{sec:appendix_rotcontrol}).

\begin{table}[t]
\centering
\footnotesize
\setlength{\tabcolsep}{5pt}
\begin{tabular}{l@{\hspace{6pt}}cccc}
\toprule
Task & Chance & Base & ZS-CoT $\Delta$ & Ctx-CoT $\Delta$ \\
\midrule
Reading 2-way   & 50.0 & 66.5 & $-1.3$ & $-3.3$ \\
Reading 3-way   & 33.3 & 50.1 & $-0.6$ & $-2.5$ \\
Reading 5-way   & 20.0 & 33.9 & $-0.7$ & $-2.8$ \\
Inference 3-way & 33.3 & 35.6 & $+0.4$ & $+0.4$ \\
Validity Y/N    & 50.0 & 50.6 & $-2.2$ & $-0.1$ \\
Validity 2AFC   & 50.0 & 51.4 & $-1.0$ & $-0.4$ \\
\bottomrule
\end{tabular}
\caption{CoT ablation, mean balanced accuracy (\%). \emph{Base} is the non-CoT score, \emph{ZS-CoT} is zero-shot CoT, and \emph{Ctx-CoT} is a task-specific scaffold pointing to the relevant lighting cues (comprehension tasks only). Means are over $19$ open models, with InternVL3-8B excluded from the 2AFC for its parser floor, as in Table~\ref{tab:cot_permodel}.}
\label{tab:cot_summary}

\end{table}

\subsection{CoT does not supply the missing prior}
In Table~\ref{tab:cot_summary}, we test two scaffolds on the comprehension tasks, a generic zero-shot CoT~\cite{wei2022chain,kojima2022large} and a contextual scaffold that names the shadow as removed and enumerates the alternative cues (neighboring shadows, sky, lighting) to reason from. Neither scaffold helps, and the comprehension tasks are unmoved. On direction inference, the mean shift is $+0.4$ points for zero-shot CoT and $+0.4$ for the contextual scaffold. On validity, the shift stays within 3 points ($-2.2$ on Yes/No, $-0.1$ with the contextual scaffold, and $-1.0$ on the bias-free 2AFC, the test that removes the yes-bias). The per-model shifts are mostly small and centered near zero, with occasional larger swings (down to $-19$ points) that are inconsistent in sign across models in Appendix Table~\ref{tab:cot_permodel}, the signature of added noise rather than added competence. Even on the easy Reading tasks, CoT is a wash (mean $-1.3$ / $-0.6$ / $-0.7$). The bottleneck is not a failure to reason over a physical variable the model possesses.

\begin{table}[t]
\centering
\footnotesize
\setlength{\tabcolsep}{3.5pt}
\newcommand{\grouprow}[1]{%
  \midrule
  \addlinespace[0pt]
  \multicolumn{7}{c}{\itshape #1}\\
  \addlinespace[-3pt]
  \midrule
}
\begin{tabular}{l ccc ccc}
\toprule
 & \multicolumn{3}{c}{Direction inference} & \multicolumn{3}{c}{Physical validity}\\
\cmidrule(lr){2-4}\cmidrule(lr){5-7}
Model & Direct & $+$Gen & $\Delta$ & Direct & $+$Gen & $\Delta$\\
\grouprow{$\leq 4$B (2--4B)}
Qwen2-VL-2B    & 28.5 & 33.3 & $+4.8$  & 50.0 & 50.0 & $+0.0$ \\
Qwen3-VL-2B    & 33.0 & 33.6 & $+0.6$  & 50.0 & 65.9 & $+15.9$ \\
Qwen2.5-VL-3B  & 33.8 & 47.6 & $+13.8$ & 50.0 & 50.0 & $+0.0$ \\
Gemma-4-E4B    & 34.4 & 52.1 & $+17.7$ & 50.0 & 69.7 & $+19.7$ \\
\grouprow{7--9B}
Qwen2-VL-7B    & 41.7 & 68.5 & $+26.8$ & 51.3 & 74.3 & $+23.0$ \\
Qwen2.5-VL-7B  & 33.3 & 50.4 & $+17.1$ & 50.2 & 76.8 & $+26.6$ \\
InternVL3-8B   & 33.3 & 55.6 & $+22.3$ & 11.9$^\dagger$ & 73.3 & -- \\
Qwen3-VL-8B    & 33.5 & 63.0 & $+29.5$ & 50.2 & 75.4 & $+25.3$ \\
\grouprow{12--14B}
Gemma-4-12B    & 42.0 & 77.2 & $+35.2$ & 56.0 & 88.6 & $+32.7$ \\
Pixtral-12B    & 34.2 & 51.3 & $+17.1$ & 53.4 & 72.7 & $+19.3$ \\
InternVL3-14B  & 34.5 & 57.3 & $+22.8$ & 49.8 & 76.4 & $+26.6$ \\
\grouprow{26--38B}
Gemma-4-26B    & 43.8 & 79.7 & $+35.9$ & 51.9 & 87.1 & $+35.2$ \\
Qwen3-VL-30B   & 37.0 & 63.8 & $+26.7$ & 52.5 & 74.8 & $+22.2$ \\
Gemma-4-31B    & 42.0 & \textbf{79.7} & $\mathbf{+37.8}$ & 50.3 & \textbf{89.6} & $\mathbf{+39.3}$ \\
Qwen2.5-VL-32B & 34.2 & 56.3 & $+22.1$ & 50.0 & 80.3 & $+30.3$ \\
InternVL3-38B  & 33.6 & 61.6 & $+27.9$ & 51.4 & 78.9 & $+27.5$ \\
\grouprow{72--78B}
Qwen2.5-VL-72B & 37.5 & 71.5 & $+34.0$ & 55.2 & 69.4 & $+14.3$ \\
LLaVA-OV-72B   & 33.9 & 58.6 & $+24.7$ & 54.7 & 75.9 & $+21.2$ \\
InternVL3-78B  & 32.7 & 57.2 & $+24.5$ & 48.6 & 67.7 & $+19.2$ \\
\midrule
Mean & 35.6 & 58.9 & $+23.2$ & 51.4 & 73.5 & $+22.1$ \\
\bottomrule
\end{tabular}
\caption{\textbf{Borrowing the IGM's shadow prior recovers both comprehension tasks} (balanced accuracy \%; chance $33.3$/$50$). \emph{Direct}: the VLM alone (Infer column of Table~\ref{tab:main}; 2AFC on $I_{\mathrm{sdw}}$ vs.\ $I_{\mathrm{flip}}$). \emph{$+$Gen}: the IGM places the shadow the scene implies, and the VLM reads it (inference) or matches the candidates against it (validity). Both tasks recover well above chance for every model above the $2$--$3$B tier. $^\dagger$Parser floor on the direct 2AFC; $\Delta$ omitted.}
\label{tab:recovery}

\end{table}

\subsection{The missing knowledge already lives in image-generation models}
\label{subsec:image}
\paragraph{T2: borrowing the IGM's shadow prior.} If the inference failure is one of missing knowledge rather than missing reasoning, then supplying that knowledge from outside should let the VLM recover the direction it cannot infer on its own. We test this by routing the VLM through an IGM, a loop that the VLM drives end-to-end. For each shadow-removed scene, the VLM writes an editing prompt asking an image editing model to add a cast shadow, the editor synthesizes it, and the same VLM then \emph{reads} the direction of that synthesized shadow, the task it is good at, instead of inferring the missing one. The prompt never specifies which way the shadow should point, so the IGM must decide that itself from the same scene cues the VLM had, the background shadows, the sky, and the shading on the object. Of the four IGMs we tested, we route FLUX.1-Kontext-dev, and detail the comparison in Appendix Sec.~\ref{sec:appendix_editor_choice}.

As shown in Table~\ref{tab:recovery}, direction inference rises from $35.6\%$ to $58.9\%$, with every one of the $19$ models improving. What matters is not the size of the gain but where it comes from. The prompt and the read-back are the VLM's own, so the recovery can only come from the IGM's shadow prior, placed in the direction the scene implies from the very cues that left the VLM at chance. The IGM holds this prior as a by-product of its training, as learning to reconstruct physically plausible images forces it to place a cast shadow where the lighting demands.

\noindent
\textbf{The gain comes from the IGM's prior, not from the decomposition.} One could object that generation-then-reading, rather than the IGM's knowledge, produces the gain. The pipeline carries nothing of its own. Holding it fixed and injecting a \emph{random} direction leaves reading at chance against the scene's truth ($30.7\%$), and injecting the \emph{ground-truth} direction reaches $67.9\%$, so what the VLM recovers is exactly what was put in. FLUX, choosing the direction itself from the shadow-free scene, is read at $65.0\%$, within three points of that ceiling. It is the editor that infers where the light comes from, and three blind annotators confirm this on its output, which places the missing shadow in the GT direction on $85.9\%$ of scenes and repairs a wrong-direction one on $83.2\%$ (Appendix Secs.~\ref{sec:appendix_dirsource} and \ref{sec:appendix_humangen}).

\paragraph{T3: judging validity from a synthesized shadow.} The same prior reaches the validity failure, where even the best VLM, Gemini-3.1-pro, judges the bias-free 2AFC at only $63\%$. We let the IGM answer instead, through the same loop. The VLM writes the editing prompt, FLUX.1-Kontext-dev infers the shadow direction from the scene's lighting cues and draws the shadow that direction implies, and the VLM then picks which of $I_{\mathrm{sdw}}$/$I_{\mathrm{flip}}$ matches it, turning the judgment into the matching task the VLM is good at. The 2AFC rises from $51.4\%$ to $73.5\%$ (Table~\ref{tab:recovery}). One might worry this is circular, that the same IGM both builds and judges the wrong-direction image. It does not, as $I_{\mathrm{flip}}$ is made by the mirror of Sec.~\ref{sec:shadowbench}, never by an IGM, so the judging IGM plays no part in constructing the pair it scores.

The same borrowed prior thus recovers both comprehension tasks, confirming the gap is one of missing knowledge that IGMs already hold. The loop, however, surfaces this prior only at test time, leaning on a separate model for every query. The lasting route is to acquire it in the weights, a direction we return to in Sec.~\ref{sec:limitations}.

\section{Limitations \& Future works}
\label{sec:limitations}
\noindent
\textbf{From borrowing the prior to acquiring it.} Our pipeline routes a single editor through every VLM and is an existence proof rather than a tuned system, so its gains are a lower bound. Acquiring the prior in the weights, rather than borrowing it at test time, is future work, by folding generation or reconstruction objectives into VLM training. Whether such a reconstructive objective is a general route to physical understanding remains open.

\noindent
\textbf{Scope: one probe of illumination, not physics in general.} ShadowBench measures a single relation, the screen-space direction of a cast shadow relative to its scene, and our claims are limited to it. Shading, specular and inter-reflections, soft shadows, multiple light sources, and material effects are untested, and reading a coarse screen-space direction is not equivalent to recovering a physical light source. Our explanation is itself a training-objective account, as image-conditioned text never requires the shadow--illumination relation while reconstruction does, so the failure reflects what current training instills rather than an inherent inability. A single cast shadow is only the first step, yet the same account extends naturally to the other cues, which is why transferring the objective, rather than more prompting, is the route we advocate.

\noindent
\textbf{Continuous-angle evaluation.} The reading and inference tasks discretize the continuous shadow-direction space into two-, three-, and five-way bins. The ground truth is annotated as continuous arrows. The discrete bins are an evaluation choice, the easiest version of the task, at which models already fail. In a preliminary finer-grained probe, we ask models to report the visible shadow's direction on a $12$-hour clock face ($30^\circ$ bins). The best model reaches $31\%$ exact and $44\%$ within $\pm 30^\circ$ (chance $8.3\%$/$25\%$; five-model means $16\%$/$32\%$), so directional emission degrades sharply well before the continuous limit. A regression formulation awaits a reliable interface for continuous angles and is future work that the continuous annotations already support.

\section{Conclusion}
\label{sec:conclusion}
We presented ShadowBench, which asks one physical question: in which direction does a cast shadow extend, in three increasing-difficulty forms. Reading a visible shadow is easy and scales with model size; the best model achieves $92\%$ against humans' $98\%$. Inferring the same direction once the shadow is removed collapses to near chance for every model, the single frontier exception reaching only $51\%$ against humans’ $86\%$, and judging physical validity is at chance for all but one model ($63\%$ vs.\ human $91\%$). CoT closes neither gap. The dissociation between reading and inference shows the failure is a lack of knowledge, not reasoning. That knowledge is not absent from the multimodal stack. Image generation models (IGMs) must place a plausible shadow whenever they synthesize an object on a ground plane, and routing one editor's output back through the VLM's strong reading ability lifts inference by $+23$ points on average. The missing cast-shadow direction prior already lives in IGMs. ShadowBench isolates this gap, and the generation-grounded recovery points toward closing it, a first step toward illumination-aware multimodal models.

\section*{Acknowledgments}
This work was supported by the BMBF Project Albatross (Grant 01IW24002). All compute was done thanks to the Pegasus cluster at DFKI.

\bibliographystyle{ieeenat_fullname}
\bibliography{main}

\maketitlesupplementary
\setcounter{page}{1}

\section{Reading is the same on the capture and the composite}
\label{sec:appendix_readsource}
Direction reading uses the composite $I_{\mathrm{sdw}}$ rather than the real capture $I_{\mathrm{all}}$, to share the compositing pipeline of the other tasks. $I_{\mathrm{sdw}}$ re-inserts the object's own extracted shadow onto the shared background, so it differs from $I_{\mathrm{all}}$ only by faint residuals left by the compositing. To check that this does not change what the models read, we run the three-way reading task on both $I_{\mathrm{all}}$ and $I_{\mathrm{sdw}}$ for each of the $19$ open models over the same $458$ scenes. The gap is negligible. As shown in Table~\ref{tab:readsource}, the mean balanced accuracy is $60.1\%$ on $I_{\mathrm{all}}$ and $61.2\%$ on $I_{\mathrm{sdw}}$, a $+1.1$-point shift that, if anything, slightly favors the cleaner composite, and every model stays within $3$ points. Reading accuracy is thus a property of the shadow and its direction, not of whether the pixels come from the raw photograph or the recomposite, so the reading ceiling reported in the main text is unaffected by the choice of source.

\section{Shadow perception scales with opacity}
\label{sec:appendix_sweep}
A precondition for our reading and inference results is that the VLMs actually \emph{perceive} the shadow when it is present. Otherwise, the failures would reduce to a perception problem rather than a reasoning one. We test this directly with an opacity sweep. Starting from the shadow-removed plate, we recompose the object's own cast shadow at a graded set of opacities from $0$ (no shadow) to $1$ (full strength), and at each level ask the single-image detection question, \emph{does the main object cast a visible shadow?} If the models perceive the shadow, the detection rate should rise smoothly with opacity.

\begin{figure}[t]
\centering
\includegraphics[width=0.92\columnwidth]{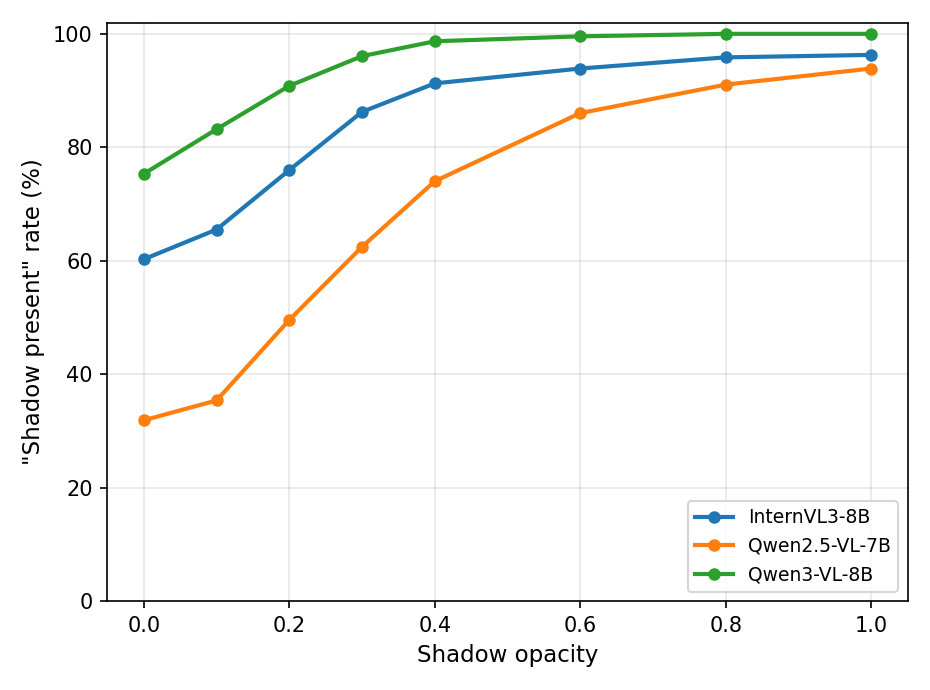}
\caption{Shadow-detection psychometric curve. The object's own cast shadow is recomposed at opacities from $0$ (no shadow) to $1$ (full), and at each level we ask whether the object casts a visible shadow. For the three models shown (InternVL3-8B, Qwen2.5-VL-7B, Qwen3-VL-8B), the detection rate rises smoothly and monotonically with opacity, showing that shadow presence is perceived as a graded quantity that scales with physical strength. Some models instead answer ``yes'' at every opacity, a response bias rather than a perceptual deficit.}
\label{fig:opacity_sweep}
\end{figure}

Figure~\ref{fig:opacity_sweep} shows that the ``shadow present'' rate climbs monotonically and smoothly with opacity, a clean psychometric curve (Qwen2.5-VL-7B $31.9 \!\rightarrow\! 93.9\%$, InternVL3-8B $60.3 \!\rightarrow\! 96.3\%$, Qwen3-VL-8B $75.3 \!\rightarrow\! 100\%$). The graded response confirms that the shadow is registered as a perceptual quantity whose salience scales with its physical strength. Shadow \emph{presence} is thus reliably available to the models, so the comprehension failures cannot be attributed to a perceptual deficit.

\begin{table}[t]
\centering
\footnotesize
\setlength{\tabcolsep}{6pt}
\begin{tabular}{l@{\hspace{8pt}}ccc}
\toprule
Model & $I_{\mathrm{all}}$ & $I_{\mathrm{sdw}}$ & $\Delta$ \\
\midrule
Qwen2-VL-2B     & 33.3 & 33.3 & $+0.0$ \\
Qwen3-VL-2B     & 33.3 & 35.7 & $+2.4$ \\
Qwen2.5-VL-3B   & 46.7 & 47.2 & $+0.5$ \\
Gemma-4-E4B     & 59.1 & 61.8 & $+2.7$ \\
Qwen2-VL-7B     & 73.2 & 71.6 & $-1.6$ \\
Qwen2.5-VL-7B   & 49.8 & 48.1 & $-1.7$ \\
InternVL3-8B    & 52.9 & 55.9 & $+3.0$ \\
Qwen3-VL-8B     & 66.2 & 66.7 & $+0.5$ \\
Gemma-4-12B     & 83.4 & 85.8 & $+2.3$ \\
Pixtral-12B     & 42.7 & 45.6 & $+2.9$ \\
InternVL3-14B   & 57.9 & 60.4 & $+2.4$ \\
Gemma-4-26B     & 87.6 & 86.6 & $-1.1$ \\
Qwen3-VL-30B    & 62.1 & 63.1 & $+1.0$ \\
Gemma-4-31B     & 84.5 & 86.8 & $+2.3$ \\
Qwen2.5-VL-32B  & 58.2 & 57.9 & $-0.3$ \\
InternVL3-38B   & 58.3 & 58.4 & $+0.1$ \\
Qwen2.5-VL-72B  & 74.8 & 75.8 & $+1.0$ \\
LLaVA-OV-72B    & 60.8 & 62.3 & $+1.6$ \\
InternVL3-78B   & 56.5 & 59.2 & $+2.7$ \\
\midrule
Mean & 60.1 & 61.2 & $+1.1$ \\
\bottomrule
\end{tabular}
\caption{\textbf{Reading is the same on the capture $I_{\mathrm{all}}$ and the composite $I_{\mathrm{sdw}}$.} Three-way direction reading on each image source and their difference $\Delta$. The means differ by $+1.1$ points, so the reading ceiling is unchanged by the choice of source.}
\label{tab:readsource}
\end{table}

\section{Which lighting conditions are hardest?}
\label{sec:appendix_shadowlen}
Having varied the shadow's strength, we now ask whether its \emph{length} affects reading and inference. We measure $\text{shadow reach}/\text{object height}$ for each of the $458$ scenes (the $95$th-percentile distance of the shadow mask from the object centroid, normalized by object height) and split at the median ratio ($1.15$) into short- and long-shadow halves. Table~\ref{tab:shadow_length} shows the mean accuracy in each half. The trend is modest but monotonic. Reading improves with shadow length (three-way $61.0\!\rightarrow\!64.4\%$ from short to long), and inference is weakest on short shadows, the output sitting at $36.5\%$, near the $33\%$ chance level, and rising to $41.9\%$ for long shadows. The hardest case is thus a short cast shadow with little geometry to anchor the direction. The effect is consistent across tasks, but small, so directional understanding is limited across the board rather than confined to a single shadow-length regime.

\begin{table}[t]
\centering
\footnotesize
\setlength{\tabcolsep}{6pt}
\begin{tabular}{l@{\hspace{10pt}}cc@{\hspace{10pt}}c}
\toprule
& \multicolumn{2}{c}{Shadow length} & \\
\cmidrule(lr){2-3}
Task & short & long & chance \\
& {\scriptsize$<1.15$} & {\scriptsize$>1.15$} & \\
\midrule
Reading 2-way   & 73.9 & \textbf{75.1} & 50.0 \\
Reading 3-way   & 61.0 & \textbf{64.4} & 33.3 \\
Inference 3-way & 36.5 & \textbf{41.9} & 33.3 \\
\bottomrule
\end{tabular}
\caption{Reading and inference accuracy stratified by shadow length, split at the median $\text{shadow reach}/\text{object height}$ ($1.15$), the mean over the $19$ open models, $229$ scenes per half. Every task improves monotonically with shadow length, and inference on short-shadow scenes sits near the $33\%$ chance level. \textbf{Bold} marks the better half per row.}
\label{tab:shadow_length}
\end{table}

\section{Presence vs.\ direction validity (per model)}
\label{sec:appendix_presence}
Table~\ref{tab:presence} gives the per-model breakdown behind the presence vs.\ direction comparison of Sec.~\ref{sec:results}. The presence 2AFC pits the correct shadow against the shadow-removed image, the direction 2AFC against the wrong-direction one. Every open model above the $2$--$3$B tier is far above chance on presence yet at chance on direction, and the gap $\Delta$ localizes the failure to shadow \emph{direction} rather than shadow presence.

\begin{table}[t]
\centering
\footnotesize
\setlength{\tabcolsep}{5pt}
\begin{tabular}{l ccc}
\toprule
 & Presence & Direction & \\
Model & 2AFC & 2AFC & $\Delta$ \\
\midrule
Qwen2-VL-2B    & 58.0 & 50.0 & $+8.0$  \\
Qwen3-VL-2B    & 51.9 & 50.0 & $+1.9$  \\
Qwen2.5-VL-3B  & 54.7 & 50.0 & $+4.7$  \\
Gemma-4-E4B    & 85.0 & 50.0 & $+35.0$ \\
Qwen2-VL-7B    & 84.6 & 51.3 & $+33.3$ \\
Qwen2.5-VL-7B  & 84.8 & 50.2 & $+34.6$ \\
InternVL3-8B   & 65.7 & $11.9^\dagger$ & $-$ \\
Qwen3-VL-8B    & 69.9 & 50.2 & $+19.7$ \\
Gemma-4-12B    & \textbf{93.3} & 56.0 & $+37.3$ \\
Pixtral-12B    & 87.5 & 53.4 & $+34.1$ \\
InternVL3-14B  & 61.1 & 49.8 & $+11.3$ \\
Gemma-4-26B    & 92.8 & 51.9 & $+40.9$ \\
Qwen3-VL-30B   & 80.1 & 52.5 & $+27.6$ \\
Gemma-4-31B    & 92.0 & 50.3 & $+41.7$ \\
Qwen2.5-VL-32B & 86.7 & 50.0 & $+36.7$ \\
InternVL3-38B  & 84.8 & 51.4 & $+33.4$ \\
Qwen2.5-VL-72B & \textbf{93.3} & 55.2 & $+38.1$ \\
LLaVA-OV-72B   & 92.3 & 54.7 & $+37.6$ \\
InternVL3-78B  & 84.8 & 48.6 & $+36.2$ \\
\midrule
\textit{Mean} & \textit{79.9} & \textit{51.4} & \textit{$+28.5$} \\
\textit{Chance}     & 50.0 & 50.0 & \\
\bottomrule
\end{tabular}
\caption{\textbf{Presence vs.\ direction validity 2AFC} on the $458$ validity scenes ($916$ counterbalanced records each). \emph{Presence} pits the correct-shadow image against the same object casting \emph{no} shadow. \emph{Direction} is the main-text validity 2AFC, pitting the correct shadow against a wrong-\emph{direction} one. Both are counterbalanced, so a constant side cancels to the $50\%$ chance. Every open model above the $2$--$3$B tier is far above chance on presence yet at chance on direction. The gap $\Delta$ isolates shadow \emph{direction}, not shadow presence, as the locus of the validity failure. $\dagger$: InternVL3-8B returns a verbose string rather than a single letter on the direction 2AFC, so that score is a parser floor and its $\Delta$ is omitted. This means we exclude InternVL3-8B, as in Tables~\ref{tab:cot_summary} and \ref{tab:recovery}, and the mean $\Delta$ is the difference between the two column means.}
\label{tab:presence}
\end{table}

\section{Per-class breakdown, full panel}
\label{sec:appendix_perclass}
Table~\ref{tab:perclass_full} extends the per-class breakdown of Table~\ref{tab:perclass} to the $19$ open VLMs and the two proprietary models, on the three comprehension tasks. The pattern is uniform. On inference, $14$ of the $19$ open models commit at least $80\%$ of their predictions to a single direction. On the single-image Yes/No task, every model collapses to a single answer: $16$ to ``yes'' and the larger Gemma models and Gemini to ``no''. On the 2AFC, most pins point to one side. The balanced score near chance is therefore a constant, image-independent answer, not per-image guessing. Gemini-3.1-pro is the only model whose predictions spread near the label priors and whose balanced score clears chance on the comprehension tasks.

\begin{table*}[t]
\centering
\footnotesize
\setlength{\tabcolsep}{5pt}
\begin{tabular}{l cccc cc ccc}
\toprule
 & \multicolumn{4}{c}{Direction 3-way} & \multicolumn{2}{c}{Validity Y/N} & \multicolumn{3}{c}{Validity 2AFC} \\
\cmidrule(lr){2-5}\cmidrule(lr){6-7}\cmidrule(lr){8-10}
Model & Read & Infer & rec L/D/R & top & bal & top & bal & top & both$\checkmark$ \\
\midrule
Qwen2-VL-2B    & 33.3 & 28.5 & 0/67/18 & D\,73 & 50.0 & Y\,100 & 50.0 & A\,100 & 0.0 \\
Qwen3-VL-2B    & 35.7 & 33.0 & 0/0/99 & R\,100 & 50.0 & Y\,100 & 50.0 & B\,100 & 0.0 \\
Qwen2.5-VL-3B  & 47.2 & 33.8 & 1/1/100 & R\,100 & 50.3 & Y\,100 & 50.0 & A\,100 & 0.3 \\
Gemma-4-E4B    & 61.8 & 34.4 & 0/5/98 & R\,96 & 49.7 & Y\,62 & 50.0 & B\,100 & 0.0 \\
Qwen2-VL-7B    & 71.6 & 41.7 & 9/59/57 & R\,51 & 50.6 & Y\,98 & 51.3 & A\,84 & 9.4 \\
Qwen2.5-VL-7B  & 48.1 & 33.3 & 0/0/100 & R\,100 & 51.1 & Y\,96 & 50.2 & B\,99 & 0.6 \\
InternVL3-8B   & 55.9 & 33.3 & 0/0/100 & R\,100 & 50.0 & Y\,100 & 11.9$^\dagger$ & -- & -- \\
Qwen3-VL-8B    & 66.7 & 33.5 & 5/0/95 & R\,96 & 50.2 & Y\,100 & 50.2 & B\,100 & 0.3 \\
Gemma-4-12B    & 85.8 & 42.0 & 30/13/83 & R\,75 & 50.0 & N\,100 & 56.0 & B\,52 & 29.2 \\
Pixtral-12B    & 45.6 & 34.2 & 0/5/98 & R\,97 & 50.6 & Y\,96 & 53.4 & A\,70 & 13.8 \\
InternVL3-14B  & 60.4 & 34.5 & 14/0/90 & R\,86 & 50.2 & Y\,100 & 49.8 & B\,100 & 0.0 \\
Gemma-4-26B    & 86.6 & 43.8 & 10/28/93 & R\,84 & 52.5 & N\,90 & 51.9 & B\,68 & 18.8 \\
Qwen3-VL-30B   & 63.1 & 37.0 & 90/0/21 & L\,81 & 51.4 & Y\,94 & 52.5 & B\,93 & 9.4 \\
Gemma-4-31B    & 86.8 & 42.0 & 16/15/94 & R\,87 & 51.3 & N\,50 & 50.3 & B\,98 & 2.2 \\
Qwen2.5-VL-32B & 57.9 & 34.2 & 4/0/99 & R\,98 & 50.3 & Y\,99 & 50.0 & A\,100 & 0.3 \\
InternVL3-38B  & 58.4 & 33.6 & 3/0/98 & R\,98 & 51.6 & Y\,76 & 51.4 & B\,93 & 7.5 \\
Qwen2.5-VL-72B & 75.8 & 37.5 & 42/0/71 & R\,66 & 50.8 & Y\,97 & 55.2 & B\,61 & 23.8 \\
LLaVA-OV-72B   & 62.3 & 33.9 & 14/1/87 & R\,84 & 51.3 & Y\,97 & 54.7 & A\,69 & 20.7 \\
InternVL3-78B  & 59.2 & 32.7 & 40/0/58 & R\,59 & 50.5 & Y\,95 & 48.6 & A\,70 & 16.0 \\
\midrule
GPT-5          & 81.7 & 46.4 & 46/20/74 & R\,57 & 52.7 & Y\,92 & 54.1 & B\,85 & 17.2 \\
Gemini-3.1-pro & 88.6 & 50.8 & 59/30/63 & R\,48 & 52.4 & N\,77 & 63.0 & A\,61 & 45.5 \\
\bottomrule
\end{tabular}
\caption{\textbf{The read-to-infer dissociation, per class, full panel.} Columns as in Table~\ref{tab:perclass}: \emph{Read}/\emph{Infer} are three-way direction balanced accuracy with the shadow present/removed, \emph{rec L/D/R} the per-direction inference recall, and \emph{top} the single most-predicted class with the percentage of items assigned to it. Open models are listed by size, the two proprietary models below. Reading scales with size while inference stays near the $33.3\%$ chance for all but Gemini-3.1-pro, and the gap is largest for the strongest readers; the single-image validity and the 2AFC likewise collapse onto a fixed class. $^\dagger$InternVL3-8B emits a verbose string rather than a single letter on the 2AFC, a parser floor, so its top class is undefined. \emph{both$\checkmark$}: share of pairs answered correctly under both presentation orders of the 2AFC; a fixed-side strategy scores $0\%$ and random guessing $25\%$, so only Gemini-3.1-pro clearly exceeds the random reference.}
\label{tab:perclass_full}
\end{table*}

\section{Per-model CoT breakdown}
\label{sec:appendix_cot}
Table~\ref{tab:cot_permodel} reports the full CoT results behind the means of Table~\ref{tab:cot_summary}, for the three comprehension tasks (direction inference, validity Yes/No, and the bias-free validity 2AFC). For each model, we list the non-CoT base score and the change under generic zero-shot CoT (ZS-$\Delta$) and the task-specific contextual scaffold (Ctx-$\Delta$). The deltas are sizable for individual models but inconsistent in sign and centered near zero, with no model lifted from chance to competence on any comprehension task. This is the per-model spread that the means in Table~\ref{tab:cot_summary} average over.

\begin{table*}[t]
\centering\footnotesize\setlength{\tabcolsep}{4pt}
\begin{tabular}{l ccc ccc ccc}
\toprule
 & \multicolumn{3}{c}{Inference 3-way} & \multicolumn{3}{c}{Validity Y/N} & \multicolumn{3}{c}{Validity 2AFC}\\
\cmidrule(lr){2-4}\cmidrule(lr){5-7}\cmidrule(lr){8-10}
Model & Base & ZS-$\Delta$ & Ctx-$\Delta$ & Base & ZS-$\Delta$ & Ctx-$\Delta$ & Base & ZS-$\Delta$ & Ctx-$\Delta$\\
\midrule
Qwen2-VL-2B & 28.5 & $+2.5$ & $+4.8$ & 50.0 & $+1.1$ & $-1.9$ & 50.0 & $+0.0$ & $+0.0$ \\
Qwen3-VL-2B & 33.0 & $-1.1$ & $+0.4$ & 50.0 & $-9.1$ & $+0.0$ & 50.0 & $+0.2$ & $+1.2$ \\
Qwen2.5-VL-3B & 33.8 & $-1.0$ & $+3.6$ & 50.3 & $+0.5$ & $-0.3$ & 50.0 & $-0.9$ & $+0.0$ \\
Gemma-4-E4B & 34.4 & $+0.3$ & $+2.1$ & 49.7 & $-0.6$ & $+1.0$ & 50.0 & $+0.0$ & $+0.0$ \\
Qwen2-VL-7B & 41.7 & $-4.5$ & $-7.6$ & 50.6 & $-0.2$ & $+1.7$ & 51.3 & $+0.0$ & $+0.5$ \\
Qwen2.5-VL-7B & 33.3 & $+0.1$ & $+0.0$ & 51.1 & $+0.2$ & $+1.9$ & 50.2 & $+0.3$ & $-1.4$ \\
InternVL3-8B & 33.3 & $+0.8$ & $-1.1$ & 50.0 & $-1.7$ & $+2.4$ & 11.9$^\dagger$ & -- & -- \\
Qwen3-VL-8B & 33.5 & $+4.0$ & $+2.8$ & 50.2 & $-0.2$ & $-0.2$ & 50.2 & $+0.6$ & $+1.3$ \\
Gemma-4-12B & 42.0 & $+1.6$ & $+7.5$ & 50.0 & $-6.4$ & $-2.0$ & 56.0 & $-2.9$ & $-2.3$ \\
Pixtral-12B & 34.2 & $-0.3$ & $-0.8$ & 50.6 & $+0.5$ & $-0.3$ & 53.4 & $-3.3$ & $-1.9$ \\
InternVL3-14B & 34.5 & $+1.1$ & $+0.9$ & 50.2 & $-0.3$ & $+0.9$ & 49.8 & $-0.9$ & $+0.6$ \\
Gemma-4-26B & 43.8 & $-4.6$ & $-4.1$ & 52.5 & $-4.2$ & $-2.5$ & 51.9 & $-3.4$ & $-3.6$ \\
Qwen3-VL-30B & 37.0 & $-2.3$ & $-3.7$ & 51.4 & $-18.9$ & $-0.8$ & 52.5 & $-2.5$ & $-1.6$ \\
Gemma-4-31B & 42.0 & $+0.9$ & $+0.1$ & 51.3 & $+3.1$ & $+1.0$ & 50.3 & $-0.1$ & $+3.9$ \\
Qwen2.5-VL-32B & 34.2 & $+3.9$ & $+1.0$ & 50.3 & $-2.4$ & $-0.3$ & 50.0 & $+0.5$ & $+1.1$ \\
InternVL3-38B & 33.6 & $+3.2$ & $+0.7$ & 51.6 & $-1.9$ & $-1.6$ & 51.4 & $-0.4$ & $-1.2$ \\
Qwen2.5-VL-72B & 37.5 & $+0.3$ & $-1.0$ & 50.8 & $-0.8$ & $-0.9$ & 55.2 & $-3.7$ & $-3.6$ \\
LLaVA-OV-72B & 33.9 & $-0.5$ & $+2.4$ & 51.3 & $+0.0$ & $-0.6$ & 54.7 & $-5.0$ & $-3.1$ \\
InternVL3-78B & 32.7 & $+3.3$ & $-0.4$ & 50.5 & $-0.2$ & $+0.6$ & 48.6 & $+3.9$ & $+3.8$ \\
\bottomrule
\end{tabular}
\caption{Per-model CoT ablation on the three comprehension tasks, balanced accuracy (\%). \emph{Base} is the non-CoT score; \emph{ZS-$\Delta$} and \emph{Ctx-$\Delta$} are the change under generic zero-shot CoT and the task-specific contextual scaffold. Deltas are large in magnitude but sign-inconsistent across models and centered near zero, the signature of added noise rather than added competence; this is the per-model spread summarized in Table~\ref{tab:cot_summary}. $^\dagger$InternVL3-8B emits no parseable A/B on the 2AFC without CoT (base is a format artifact, not a judgment), so its deltas are omitted and it is excluded from the 2AFC column statistics.}
\label{tab:cot_permodel}
\end{table*}

\section{Which IGM to route through the pipeline}
\label{sec:appendix_editor_choice}
The generation-grounded pipeline in Sec.~\ref{subsec:image} routes one image editor, FLUX.1-Kontext-dev, through the VLM. Here, we justify that choice and show that the shadow-direction prior it yields is not specific to a single IGM. We evaluate all four IGMs through the identical VLM-driven loop of Sec.~\ref{subsec:image}, never a fixed instruction. For each of the $458$ scenes, the VLM writes its own editing prompt asking for a cast shadow without ever stating which way it should point, and the IGM must infer the direction from the same scene cues the VLM had. We compare the open editors FLUX.1-Kontext-dev and OmniGen2 and the proprietary gpt-image-2 and Gemini-3-pro-image, each receiving the VLM-authored prompt for the shadow-removed input. We then score each editor exactly as the pipeline uses it: we run the full read-back step, marking the object and passing the edit to a single fixed VLM reader (Gemma-4-31B, the strongest reader in Table~\ref{tab:recovery}) that reports the three-way direction (left/down/right) of the added shadow, and compare to the per-scene human ground truth. Fixing the reader across all four editors isolates the editor's prior from the reader, so the numbers are directly comparable across IGMs and to the $+$Gen column of Table~\ref{tab:recovery} (the FLUX entry is that same pipeline). Table~\ref{tab:editor_choice} reports the resulting three-way direction-inference accuracy. Two points follow. First, the prior is broadly shared. All four IGMs, open and proprietary, drive the reader far above the $33.3\%$ three-way chance ($61$--$80\%$). The shadow-direction knowledge the VLMs lack on their own is therefore a general property of modern IGMs rather than one model's quirk. Second, FLUX.1-Kontext-dev ranks highest on the three-way score ($79.7\%$), edging the proprietary Gemini-3-pro-image ($79.0\%$) while being open and freely reproducible, so we route it through the pipeline as the best reproducible choice. OmniGen2, the other open editor, trails, and the two proprietary editors regenerate larger regions of the image, further favoring a cleanly localized open editor whose edit isolates the added shadow.

\begin{table}[t]
\centering
\footnotesize
\setlength{\tabcolsep}{6pt}
\begin{tabular}{l@{\hspace{4pt}}l@{\hspace{8pt}}c}
\toprule
Editor & Type & 3-way \\
\midrule
FLUX.1-Kontext-dev  & open  & \textbf{79.7} \\
OmniGen2            & open  & 61.0 \\
gpt-image-2         & prop. & 73.4 \\
Gemini-3-pro-image  & prop. & 79.0 \\
\midrule
Chance              &       & 33.3 \\
\bottomrule
\end{tabular}
\caption{Editor-intrinsic shadow-direction accuracy on the $458$ scenes, scored through the same VLM-driven loop as the main text. Each IGM receives the VLM-authored editing prompt, which never states the shadow direction, and adds a cast shadow to the shadow-removed input; a single fixed VLM reader (Gemma-4-31B) then reports the added shadow's three-way direction (left/down/right), which we compare to the human GT. This is the same loop, metric, and reader as the $+$Gen inference column of Table~\ref{tab:recovery} (chance $33.3$); the FLUX entry is that same pipeline. \emph{prop.}\ is proprietary; Gemini-3-pro-image is nano-banana-pro. \textbf{Bold} is the best. All four IGMs exceed three-way chance by a wide margin; FLUX.1-Kontext-dev is best overall, edging the proprietary Gemini-3-pro-image while remaining open and reproducible, so we route it through the pipeline.}
\label{tab:editor_choice}
\end{table}

\section{The shadow-only rotation control}
\label{sec:appendix_rotcontrol}
This section details the control summarized in Sec.~\ref{sec:results}: validity negatives that alter \emph{only} the shadow. 

\noindent
\textbf{Construction.} A ground-cast shadow can extend slightly above the object when the light is near the horizon, but it never points straight up. We therefore exclude only a narrow cone about the vertical, admitting $\mathcal{A}=\{a : |a-90^\circ| > \theta\}$ with $\theta=30^\circ$ ($0^\circ$ = right, counter-clockwise, so $90^\circ$ is straight up), and rotate \emph{within} $\mathcal{A}$: a full-circle rotation would place the shadow directly above the object, a configuration the benchmark never contains and a leak of its own. For each of the $458$ validity scenes we recover the soft shadow matte $\alpha$ (Sec.~\ref{sec:shadowbench}) and construct three negatives by advancing the scene's ground-truth angle $a$ along $\mathcal{A}$ in equal steps, skipping the excluded cone, and keeping only offsets that land in a different three-way class than $a$. The matte is rotated about the annotated shadow origin, zeroed under the object mask, and recomposited as $I_{\mathrm{back}}\odot(1-\alpha_{\mathrm{rot}})$ with the \emph{original} object pixels pasted on top, bit-identical to the positive. The positive of each pair is the $0^\circ$ recomposite of the same pipeline, so both candidates share one processing history. Each pair is evaluated in both presentation orders, as in the main 2AFC.

\noindent \textbf{Why the control leaks non-lighting cues.} Rotating only the shadow necessarily produces an image whose shadow could not be cast by the pictured object under \emph{any} light source: the silhouette no longer matches the object's projection, the rotation can strand the shadow on implausible supports or send it behind the object, and image-space rotation breaks fine compositing statistics. Two of these are directly screenable, and we apply the screens in turn. First, the negative's visible shadow mass -- the total opacity falling on the background outside the object -- must stay within $80\%$ of the positive's, since a rotation that hides the shadow behind the object or pushes it off the plate turns the pair into a test of shadow \emph{presence}, a cue the main text shows models exploit at $80\%$; this retains $200$ of the $1{,}374$ rotated composites. Second, a human screener must rate the compositing clean, retaining $112$. That $112$ of $1{,}374$ survive both screens is itself the measurement: the overwhelming majority of shadow-only rotations are distinguishable from the positive without ever consulting the scene's lighting. The silhouette mismatch admits no such screen, as it is intrinsic to any shadow-only rotation.

\noindent \textbf{Result: accuracy tracks the leaks, not the lighting.} Table~\ref{tab:rotcontrol} reports open-panel accuracy under successively stricter curation. Uncurated, the panel reads the rotated negatives at $64.0\%$ and the five best models at $76.8\%$; equating visible shadow mass lowers these to $60.7\%/69.2\%$, and additionally requiring clean compositing to $57.1\%/62.4\%$. Accuracy falls monotonically as each non-lighting cue is removed, which is the signature of a construction whose detectability comes from the manipulation rather than from the lighting it is meant to probe, and the residual above chance is what the silhouette mismatch, which no screen can remove, still supplies. Humans answer the same curated pairs at $94.4\%$ (three annotators, $93.7/98.5/91.0\%$; majority vote $98.5\%$; mean pairwise agreement $90.0\%$; $92.9\%$ on the mass-equated subset), so the pairs remain fully decidable from lighting alone, and the models' decline is not a loss of signal but a loss of shortcut. The mirror construction avoids every one of these leaks by design, which is why the main benchmark uses it.

\begin{figure*}[t]
\centering
\includegraphics[width=\textwidth]{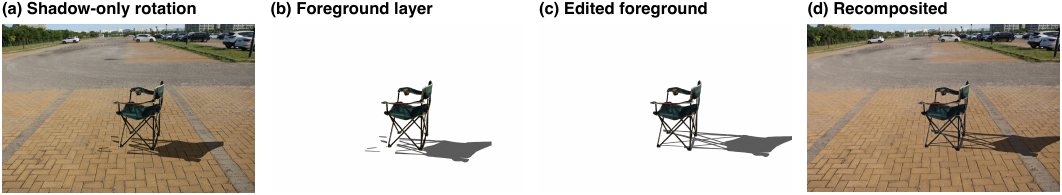}
\caption{\textbf{Naturalising the shadow-only rotation with an image editor.}
(a)~the negative a bare rotation produces: the soft shadow matte is rotated about the annotated
origin and recomposited, the object pixels left untouched. An image-space rotation cannot yield a
silhouette the pictured object could actually cast, and it disturbs the fine compositing statistics
along the shadow's edge -- cues a model can exploit without consulting the scene's lighting at all.
We therefore rebuild every negative: (b)~the object and its rotated shadow are lifted onto a white
canvas, the scene removed so the editor has no lighting context to re-infer the direction from;
(c)~FLUX.1-Kontext-dev redraws the shadow's shape under an instruction to keep its direction, length
and size, and the result is clamped to the seed's rotated support so the direction cannot drift;
(d)~the redrawn shadow is recomposited into the captured background plate, with the object pixels
taken unchanged from the positive so that the shadow is the only difference within a pair.
Detectability falls from $64.0\%$ on (a) to $56.3\%$ on (d), while humans answer the same pairs at
$94.4\%$ (Table~\ref{tab:rotcontrol}).}
\label{fig:editorchain}
\end{figure*}

\noindent
\textbf{An editor-naturalized variant converges to the same conclusion.} Because rotation alone breaks compositing, we also rebuilt every negative with an image editor (Fig.~\ref{fig:editorchain}): the rotated matte is pasted on a white canvas with the scene removed, so the editor has no lighting context to re-infer the direction from; FLUX.1-Kontext-dev redraws the shadow's shape, and the re-extracted shadow is composited back, the $0^\circ$ positive passing through the identical pipeline. Three annotators screen all $1{,}374$ edits and judge $98$ free of visible artifacts. On these, the panel reads the negatives at $52.3\%$, close to the $51.4\%$ it reaches on the mirror, and any artifact the editor leaves behind would only make the negative easier to spot, so this is an upper bound on what the manipulation leaks. Both constructions converge: removing non-lighting leaks, whether by human curation ($64.0\!\to\!57.1\%$) or by generative refinement ($64.0\!\to\!52.3\%$), drives detectability toward chance, while the leak-free mirror sits at chance.

\begin{table}[t]
\centering
\footnotesize
\setlength{\tabcolsep}{6pt}
\begin{tabular}{l cc}
\toprule
Curation level & Pooled panel & Top-5 models \\
\midrule
None (1{,}374 pairs)             & 64.0 & 76.8 \\
Shadow mass $\geq\!0.8$ (200)    & 60.7 & 69.2 \\
$+$ clean compositing (112)      & 57.1 & 62.4 \\
\midrule
Editor-refined (1{,}374)          & 56.3 & 63.8 \\
$+$ shadow mass $\geq\!0.8$ (358) & 54.6 & 58.3 \\
$+$ artifact-free (98)            & 52.3 & 55.1 \\
\midrule
Mirror (main, 458)                & 51.4 & 53.4 \\
\bottomrule
\end{tabular}
\caption{\textbf{Shadow-only rotation 2AFC under successively stricter curation} (balanced accuracy \%, chance $50$; open panel of Table~\ref{tab:main}, InternVL3-8B excluded from means for its parser floor as there). \emph{Top-5} follows the five best models on each construction's uncurated set through curation. Top block: human curation of the raw rotation (equated shadow mass, then clean compositing). Middle block: the editor-refined construction, in which FLUX.1-Kontext-dev redraws the rotated shadow's shape on a white canvas (scene hidden) before recompositing; three annotators retain $98$ of $1{,}374$ edits as artifact-free. Removing the non-lighting leaks (shadow-mass asymmetry, broken compositing, residual artifacts) drives accuracy monotonically toward chance along both routes, while the mirror construction, which has none of these leaks, sits at chance.}
\label{tab:rotcontrol}
\end{table}

\section{Option-order robustness}
\label{sec:appendix_shuffle}
The reading and inference prompts list the direction words in a fixed order. Table~\ref{tab:shuffle} repeats both tasks with the option order independently shuffled per scene, for the full open panel. Reading is essentially unchanged ($61.2\!\to\!59.2\%$ mean) and inference remains at chance ($35.6\!\to\!35.6\%$): the same manipulation leaves the high task high and the chance task at chance, as a knowledge account predicts and a formatting account cannot. Task~3 needs no such rerun, as its 2AFC already presents every pair in both orders, which cancels any fixed side to 
$50\%$ and separates preference from discrimination (Sec.~\ref{sec:results}).

\begin{table}[t]
\centering
\footnotesize
\setlength{\tabcolsep}{5pt}
\begin{tabular}{l cc cc}
\toprule
 & \multicolumn{2}{c}{Reading 3-way} & \multicolumn{2}{c}{Inference 3-way} \\
\cmidrule(lr){2-3}\cmidrule(lr){4-5}
Model & fixed & shuffled & fixed & shuffled \\
\midrule
Qwen2-VL-2B     & 33.3 & 33.9 & 28.5 & 32.0 \\
Qwen3-VL-2B     & 35.7 & 34.0 & 33.0 & 33.1 \\
Qwen2.5-VL-3B   & 47.2 & 46.3 & 33.8 & 32.3 \\
Gemma-4-E4B     & 61.8 & 59.1 & 34.4 & 35.4 \\
Qwen2-VL-7B     & 71.6 & 68.4 & 41.7 & 36.4 \\
Qwen2.5-VL-7B   & 48.1 & 51.1 & 33.3 & 33.0 \\
InternVL3-8B    & 55.9 & 50.0 & 33.3 & 33.3 \\
Qwen3-VL-8B     & 66.7 & 67.1 & 33.5 & 33.5 \\
Gemma-4-12B     & 85.8 & 83.1 & 42.0 & 43.5 \\
Pixtral-12B     & 45.6 & 48.5 & 34.2 & 38.7 \\
InternVL3-14B   & 60.4 & 50.9 & 34.5 & 33.6 \\
Gemma-4-26B     & 86.6 & 85.7 & 43.8 & 44.1 \\
Qwen3-VL-30B    & 63.1 & 57.6 & 37.0 & 33.3 \\
Gemma-4-31B     & 86.8 & 85.0 & 42.0 & 43.4 \\
Qwen2.5-VL-32B  & 57.9 & 60.7 & 34.2 & 34.9 \\
InternVL3-38B   & 58.4 & 60.6 & 33.6 & 33.3 \\
Qwen2.5-VL-72B  & 75.8 & 68.2 & 37.5 & 35.3 \\
LLaVA-OV-72B    & 62.3 & 61.3 & 33.9 & 36.3 \\
InternVL3-78B   & 59.2 & 54.3 & 32.7 & 31.6 \\
\midrule
Mean            & 61.2 & 59.2 & 35.6 & 35.6 \\
\bottomrule
\end{tabular}
\caption{\textbf{Shuffling the option-word order changes no conclusion} (balanced accuracy \%). Reading and inference rerun with the direction words listed in a per-scene random order.}
\label{tab:shuffle}
\end{table}

\section{Direction-source controls for the pipeline}
\label{sec:appendix_dirsource}
Table~\ref{tab:dirsource} holds the generation-then-reading pipeline fixed and varies only where the injected direction comes from, on a $10$-model subset spanning all sizes. \emph{Random} composites the extracted shadow matte at a random rotation and asks the model to read it; scored against the scene's ground truth it sits at chance, while scored against the direction actually injected the same predictions are far above chance -- the models faithfully read whatever shadow they are given, so the pipeline transmits exactly the knowledge injected and no more. \emph{GT composite} injects the ground-truth direction through the identical compositing pipeline and bounds what a matte-composited shadow can transmit. \emph{IGM} is the same models' $+$Gen score (Table~\ref{tab:recovery}), where FLUX chooses the direction itself: it is read at the level of the ground-truth composite, so the generator supplies essentially all of the direction knowledge the pipeline can carry.

\begin{table}[t]
\centering
\footnotesize
\setlength{\tabcolsep}{1.5pt}
\begin{tabular}{l cc c c}
\toprule
 & \multicolumn{2}{c}{Random direction} & GT comp. & IGM (FLUX) \\
\cmidrule(lr){2-3}\cmidrule(lr){4-4}\cmidrule(lr){5-5}
Model & vs.\ GT & vs.\ injected & vs.\ GT & vs.\ GT \\
\midrule
Gemma-4-E4B     & 32.0 & 49.9 & 61.8 & 52.1 \\
Qwen2.5-VL-7B   & 28.8 & 41.3 & 48.1 & 50.4 \\
Qwen3-VL-8B     & 29.7 & 50.9 & 66.7 & 63.0 \\
Gemma-4-12B     & 32.4 & 67.2 & 85.8 & 77.2 \\
Pixtral-12B     & 30.7 & 38.4 & 45.6 & 51.3 \\
Gemma-4-26B     & 32.6 & 65.3 & 86.6 & 79.7 \\
Qwen3-VL-30B    & 31.2 & 50.3 & 63.1 & 63.8 \\
Gemma-4-31B     & 31.3 & 66.0 & 86.8 & 79.7 \\
InternVL3-38B   & 31.3 & 43.9 & 58.4 & 61.6 \\
Qwen2.5-VL-72B  & 26.8 & 65.9 & 75.8 & 71.5 \\
\midrule
Mean            & 30.7 & 53.9 & 67.9 & 65.0 \\
\bottomrule
\end{tabular}
\caption{\textbf{Only the injected direction carries signal} (balanced accuracy \%, chance $33.3$). Reading a composited shadow at a \emph{random} rotation is at chance against the scene's truth yet far above chance against the direction actually injected; the identical composite in the \emph{ground-truth} direction reaches $67.9\%$; the IGM-synthesized shadow (\emph{$+$Gen} of Table~\ref{tab:recovery}, same models) is read at $65.0\%$, at ground-truth-composite level.}
\label{tab:dirsource}
\end{table}

\section{Human measurement of the generator's direction} \label{sec:appendix_humangen} 
Automatic direction extraction from a synthesized image is itself noisy, so we measure the generators the same way the ground truth is defined: by human observation. Three annotators, blind to the annotated ground-truth direction, view the shadow-free original and the edited image side by side and report the three-way direction of the added shadow, stratified by ground-truth class. For FLUX.1-Kontext-dev, the read direction matches the scene's ground truth on $85.9\%$ of scenes (three annotators; majority vote $87.9\%$; mean pairwise agreement $88.9\%$; chance $33.3\%$), independently confirming the automatic read-back ($79.7\%$, Table~\ref{tab:editor_choice}). Table~\ref{tab:editor_human} summarizes the human-rated accuracy for both the synthesis and the repair setting.

\noindent \textbf{Repair.} The same prior corrects as well as synthesizes. We give the editor the wrong-direction composite and the instruction to make the shadow consistent with the scene's lighting, restricted to the scenes whose mirrored direction falls in a different three-way class than the ground truth, so that a correct repair is distinguishable from leaving the shadow as it is. The same annotators label the corrected output identically; the shadow is redrawn into the ground-truth class in $83.2\%$ of scenes (three annotators; majority vote $82.8\%$; mean pairwise agreement $87.2\%$; chance $33.3\%$). This is precisely the consistency judgment that Task~3 poses, held by the generator but not by any VLM we test.

\begin{table}[t]
\centering
\footnotesize
\setlength{\tabcolsep}{8pt}
\begin{tabular}{l c}
\toprule
Edit set & Human 3-way acc.\ (\%) \\
\midrule
Synthesis (FLUX) & 85.9 \\
Repair (FLUX) & 83.2 \\
\bottomrule
\end{tabular}
\caption{Human-rated shadow direction of the FLUX edits (chance $33.3\%$).}
\label{tab:editor_human}
\end{table}

\section{Curation statistics and record counts}
\label{sec:appendix_curation}
From the $2{,}715$ captured scenes, automated background-consistency filtering and the three-annotator review of Sec.~\ref{sec:curation} yield $458$ scenes with reliable direction ground truth (Tasks~1--2; per stage: $2{,}715 \to 1{,}098$ after the automated filter, $458$ after the usability review). All $458$ curated scenes form the Task-3 validity set ($916$ Yes/No records; $458$ pairs $\times$ $2$ orders $= 916$ 2AFC records). Ground-truth directions are annotated as continuous arrows; the mirror changes the direction by $\geq 90^\circ$ in $85.4\%$ of the arrow-annotated validity scenes (median $153^\circ$), and re-scoring the recorded predictions without the $14$ near-vertical scenes (mirror displacement $<15^\circ$) leaves the open-panel 2AFC mean unchanged ($51.4\%$) and moves Gemini-3.1-pro from $63.0$ to $63.4\%$, changing no conclusion.

\noindent
\textbf{Class definitions.} With the annotated angle $a$ in scene coordinates ($0^\circ$ = right, counter-clockwise), the two-way task labels Right for $\cos a > 0$ and Left otherwise; the three-way task labels right for $a \in (315^\circ, 45^\circ]$, left for $a \in (45^\circ, 225^\circ]$, and down otherwise; the five-way task splits the lower half-plane at $45^\circ$ boundaries (left, down-left, down, down-right, right). Upward bins are omitted because ground-cast shadows essentially never point above the horizon in the captured scenes.

\noindent
\textbf{Source filtering.} Stage 1 automatically rejects scenes whose residual $|I_{\mathrm{all}} - I_{\mathrm{back}}|$ outside the object and effect masks exceeds a fixed threshold, which removes capture pairs with camera motion, background motion, or lighting changes; stage 2 is the unanimous three-annotator review of Sec.~\ref{sec:curation}.

\noindent
\textbf{Compute cost.} All open-model evaluations run on single nodes ($1$--$4$ GPUs) of A100/H100 class; the full panel over all tasks, ablations, and controls totals roughly $2{,}800$ GPU-hours, and each IGM edit takes $5$--$25$ seconds per image. The pipeline adds one editor call and one reading call per scene at test time.

\section{Failure cases}
\label{sec:appendix_failure}
\begin{figure}[t]
\centering
\includegraphics[width=\columnwidth]{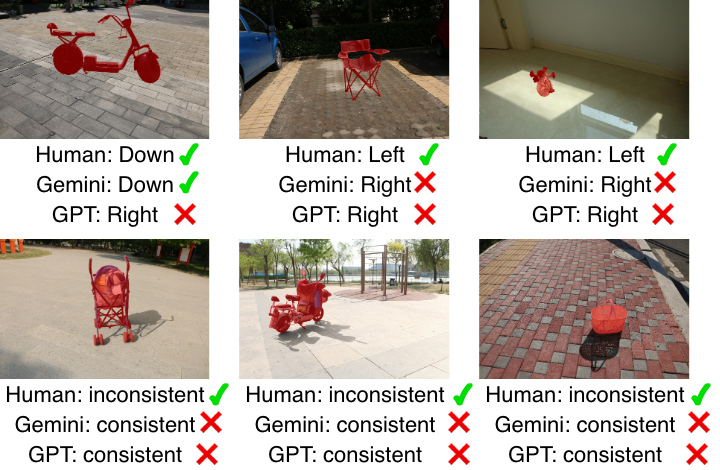}
\caption{\textbf{Failure cases.} Top: direction inference on $I_{\mathrm{obj}}$ (Task~2). Bottom: single-image validity on $I_{\mathrm{flip}}$ (Task~3). The target object is highlighted in red, as in the evaluation prompts.}
\label{fig:failure}
\end{figure}

Figure~\ref{fig:failure} shows representative failures of the two frontier models on the comprehension tasks. On inference (top), GPT-5 answers \emph{right} on every scene while Gemini-3.1-pro follows the same default on two of the three; the collapse is quantified in Table~\ref{tab:perclass_full}. On validity (bottom), both accept the mirrored composite that human annotators reject.

\end{document}